\documentclass[acmsmall,screen]{acmart}

\usepackage{epsfig}
\usepackage{graphicx}
\usepackage{amsmath}
\usepackage{epstopdf}
\usepackage{color}
\usepackage{array}
\usepackage{comment}
\usepackage{subcaption}
\usepackage{multirow}
\makeatletter  
\newif\if@restonecol  
\makeatother

\usepackage{multirow}

\AtBeginDocument{%
  \providecommand\BibTeX{{%
    \normalfont B\kern-0.5em{\scshape i\kern-0.25em b}\kern-0.8em\TeX}}}

\setcopyright{acmcopyright}
\copyrightyear{0}
\acmYear{0}
\acmDOI{0}

\acmJournal{JACM}
\acmVolume{0}
\acmNumber{0}
\acmArticle{0}
\acmMonth{0}

\acmSubmissionID{0}

\graphicspath{{./figure/}}
\newcommand{\etal}{\textit{et al.}}
\newcommand{\eg}{\textit{e.g.}}

\newcommand{\etc}{\textit{etc.}}

\newcommand{\revrm}[1]{\textcolor[rgb]{0.1,0.1,0.8}{}}

\begin{document}

\title{Neural Implicit Field Editing Considering Object-environment Interaction.}

\author{Zhihong Zeng$^{1,2,3}$}
\orcid{0009-0006-3833-4219}
\email{zengzhihong17@mails.ucas.ac.cn}

\author{Zongji Wang$^{1,2}$}
\orcid{0000-0001-9684-300X}
\email{wangzongji@aircas.ac.cn}

\author{Yuanben Zhang$^{1,2}$}
\orcid{0009-0005-9284-1919}
\authornote{Yuanben Zhang is the corresponding author. }
\email{zhangyb@aircas.ac.cn}

\author{Weinan Cai$^{1,2,3}$}
 \orcid{0009-0000-9417-2329}
 \email{caiweinan22@mails.ucas.ac.cn}

\author{Zehao Cao$^{1,2,3}$}
\email{caozehao22@mails.ucas.ac.cn}

\author{Lili Zhang$^{1,2}$}
\email{zhanglili86@126.com}

\author{Yan Guo$^{1,2}$}
\email{guoyan@aircas.ac.cn}

\author{Yanhong Zhang$^{4}$}
\email{1018553473@qq.com}

\author{Junyi Liu$^{1,2}$}
\email{liujy004735@aircas.ac.cn}

\affiliation{%
  \institution{\newline$^{1}$ Aerospace Information Research Institute, Chinese Academy of Sciences, Beijing, 100190}
  \country{China}
}

\affiliation{%
  \institution{$^{2}$ Key Laboratory of Network Information System Technology(NIST), Institute of Electronics Chinese Academy of Sciences, Beijing, 100190}
  \country{China}
}

\affiliation{%
  \institution{$^{3}$ School of Electronic, Electrical and Communication Engineering, University of Chinese Academy of Sciences, Beijing, 100190}
  \country{China}
}

\affiliation{%
  \institution{$^{4}$State Key Laboratory of Geo-information Engineering, Xi'an Research Institute of Surveying and Mapping, Xi'an}
  \country{China}
}

\renewcommand{\shortauthors}{Zeng, et al.}

\setcopyright{acmcopyright}
\acmJournal{TOMM}
\acmYear{2024} \acmVolume{0} \acmNumber{0} \acmArticle{0} \acmMonth{0} \acmPrice{0}\acmDOI{0}

\begin{abstract}
  The 3D scene editing method based on neural implicit field has gained wide attention. It has achieved excellent results in 3D editing tasks. However, existing methods often blend the interaction between objects and scene environment. The change of scene appearance like shadows is failed to be displayed in the rendering view. In this paper, we propose an Object and Scene environment Interaction aware (OSI-aware) system, which is a novel two-stream neural rendering system considering object and scene environment interaction. To obtain illuminating conditions from the mixture soup, the system successfully separates the interaction between objects and scene environment by intrinsic decomposition method. To study the corresponding changes to the scene appearance from object-level editing tasks, we introduce a depth map guided scene inpainting method and shadow rendering method by point matching strategy. Extensive experiments demonstrate that our novel pipeline produce reasonable appearance changes in scene editing tasks. It also achieve competitive performance for the rendering quality in novel-view synthesis tasks.
\end{abstract}



\maketitle

\section{Introduction} \label{sec:Introduction}

Reconstructing indoor 3D scenes in the real world and placing the 3D assets is a main development direction in the 3D computer vision field. Early 3D entities are representated as voxel occupancy~\cite{EarlyRepresentations:Occupancy1,
EarlyRepresentations:Occupancy2,
EarlyRepresentations:Occupancy3}, mesh grid~\cite{EarlyRepresentations:image2mesh,
EarlyRepresentations:pixel2mesh} or point cloud~\cite{EarlyRepresentations:dlmnet,
EarlyRepresentations:multiresolution,
EarlyRepresentations:unsupervised}, resulting in huge amount of redundant data. Novel emerging neural radiance field method~\cite{EarlyRepresentations:ImplicitField,
EarlyRepresentations:SRN} offers improvement by learning an implicit representation of a single target from a set of posed images, which dramatically ameliorates the data storage and reduces the computing resources requirement. It also provides high resolution data for 3D tasks. In order to apply the efficient representation of implicit fields to various applications, such as (VR browsing, animation CG, and simulation), scene editing tasks like moving, inserting and removing objects, changing illuminating conditions \etc, have attracted a lot of attention from researchers.

As the most popular implicit field representation method among researchers, NeRF~\cite{NeRFVariant:NeRF} represents scenes with implicit fields of volume density and view-dependent color, and achieves photo realistic novel view synthesis results. Existing scene editing methods based on NeRF have made remarkable progress, but there are still shortcomings like editing flexibility and neglect of interactions between objects and scene environment like shadows. Conditional NeRF ~\cite{NeRFVariant:ConditionalNeRF} accomplishes the editing tasks by backward parameter updating. The users' editing commands are reverse-rendered to the features that control the change of implicit field volume density $\sigma$ and color $c$. The network can then perform the corresponding editing results. However, due to the limited neural network capacity, it fails to produce more flexible editing results. Object NeRF~\cite{NeRFVariant:objectnerf} uses 3D object-level box labels and 2D masks to help implicit fields recognize object-level representations. Then it produces novel editing results by adjusting the sampled rays in the inference stage. This method fails to model the interactions between objects and scene environment, which leads to less realistic editing results.

In this paper, we offer a novel two-stream pipeline of neural rendering system, which successfully separates the interaction between objects and scene environment by intrinsic decomposition method and provides a more free and convenient solution for object-level scene editing tasks. From a set of posed images captured in real and synthesis scenes, we introduce the 3D object-level box labels and 2D masks to help implicit field generate corresponding knowledge to achieve object-level reconstruction. Our intrinsic decomposition method is adopted on 2D images to obtain albedo(reflection) and shading(illumination) map. The information is then propagated to the neural implicit field through the reverse-rendering pipeline with a new branch of the NeRF network. After that, the interaction between the objects and scene environment will be separated from the scene. Specifically, shadow information are thus recognized and separated from the mixture soup of scene appearance without additional illuminating supervision.

We explor the depth guided scene inpainting mechanism~\cite{NeRFVariant:spinnerf} and shadow rendering method by point matching strategy to utilize the separation results we obtain earlier. After that, we introduce an artificial point light source into the scene and produce object-level changes in the appearance of the scene under arbitrary illuminating conditions and objects geometry and positions. We put all the steps above together in a ``post-processing'' module to simplify the overall framework.

\textbf{Contributions.} We summarize the contributions as follows:
\vspace{-3pt}
\begin{itemize}
	\item We offer a novel OSI-aware system, which is a two-stream neural rendering system considering object and scene environment interaction. It carries out our own intrinsic decomposition method on posed images and utilizes 3D box labels and 2D masks to separate the interaction between objects and scene environment.
	
	\item We design a novel ``post-processing'' module, which utilizes the aforementioned separation results and turns them into explicit light editing components. The module simplifies the tedious explicit rendering pipeline, and achieves realistic editing results under arbitrary illuminating conditions and objects geometry and positions. 
	
	\item We demonstrate the effectiveness of the method through comprehensive experiments on multiple common datasets for neural implicit fields. Our system produces a more natural scene editing results on the datasets and achieves similar results to SoTA methods on the basic novel-view synthesis task of NeRF.
\end{itemize}

\section{Related Work}

\noindent\textbf{Neural Rendering.} 
Traditional 3D reconstruction methods encode 3D objects into different types of explicit mathematical representations, such as voxels~\cite{EarlyRepresentations:Occupancy1,
EarlyRepresentations:Occupancy2,
EarlyRepresentations:Occupancy3}, mesh grids~\cite{EarlyRepresentations:image2mesh,
EarlyRepresentations:pixel2mesh}, point clouds~\cite{EarlyRepresentations:dlmnet,
EarlyRepresentations:multiresolution,
EarlyRepresentations:unsupervised}, \etc, which are intuitive but inefficient. For example, the voxel method defines each voxel in the space range of $\mathcal{O}(n^3)$ and stores and calculates its space occupancy, which is quite data storage unfriendly and computing wasteful. However, the novel implicit field methods like SRN~\cite{EarlyRepresentations:SRN} and NeRF~\cite{NeRFVariant:NeRF} can simplify the accumulation of large amounts of data into a neural function, which not only reduce the storage and compute difficulty, but also ensure arbitrary spatial resolution for 3D representation. The classic NeRF maps spatial points to the volume density $\sigma$ and color $c$, and implement alpha compositing~\cite{NeRFTechnique:compositing} on sampled rays to predict the pixel colors. NeuS~\cite{NeRFVariant:neus} choose the signed distance~\cite{EarlyRepresentations:deepsdf} to represent a smooth surface of the target. Intrinsic NeRF~\cite{NeRFVariant:intrinsicnerf} performs unsupervised clustering of image color labels in training process to reconstruct the albedo and shading scene colors. Object NeRF~\cite{NeRFVariant:objectnerf} uses the voxel vertex coding of NSVF~\cite{NeRFVariant:NSVF} and 3D box labels in their object training branch to achieve object-level understanding of editable objects. The common feature of these NeRF variant methods is that the implicit neural network is regarded as an encoder. It controls the expression of the network parameters by adjusting the supervision, for example, the common posed image or the clustering albedo result. We can also expand the implicit field value domain, like converting the volume density $\sigma$ into a signed distance function (sdf value), or replacing rgb colors $c$ with albedo $a$. Based on this observation, we offer a novel two-stream pipeline of neural rendering network. It automatically separate the interaction between objects and scene environment because we have already preprocess the data for albedo and shading supervision for the network. Therefore, we don't need to do additional processing on the results of training to get the information we want (\eg\space performing intrinsic decomposition on the points color $c$ along the rays).

\noindent\textbf{Scene Editing.}
Implicit field has benefits in terms of precision and data storage, comparing with traditional explicit 3D editing components such as voxel and mesh grid. However, it lacks the traditional mature rendering pipeline of explicit editing methods~\cite{SceneEdit:ExplicitPipline}. We can simply resume those pipelines by query the neural field to obtain an explicit representation, but this is equivalent to abandon the implicit method's high precision and friendly storage advantages. An intuitive solution is the parameter updating method. The users' editing commands are reverse-rendered to the features that control the change of implicit field volume density $\sigma$ and color $c$, so the network can perform the corresponding editing results. Conditional NeRF~\cite{NeRFVariant:ConditionalNeRF} performs parameter updating to accomplish scene editing tasks, but the weights are changed permanently after the parameters are propagated, making it hard to roll back after the user giving commands. Due to the network capacity, this method is also not flexible which means the editing operations are limited. From Yang \etal~cite{NeRFVariant:objectnerf}, GIRAFFE~\cite{NeRFVariant:giraffe} and OSF~\cite{SceneEdit:OSFs} follow a ``bottom-up'' method, which rebuilds an implicit field for each editable object and background scene respectively, these implicit fields are then combined and supervised on their arrangement by 2D images. This method reconstructing every objects to an implicit field is not economical. It is easy to be applied to synthetic scene datasets while not in complex real scenes. Object NeRF~\cite{NeRFVariant:objectnerf} uses 3D and 2D supervision to segment editable targets objects and accomplish editing tasks by manipulating sampled rays, but it fails to focus on the interaction between objects and scene environment such as shadows. Instead, our approach separates those interaction information from the scene and turns them into explicit editing components, preserving further chance of entering the explicit rendering pipeline. It combines the intuitiveness of explicit representation and the advantages of implicit representation.

\noindent\textbf{Relighting.} 
The NeRF-based implicit field adopt a simplified absorption + luminescent rendering model~\cite{NeRFTechnique:optical}, which assumes the particles in space are self-luminous and no direct or indirect illumination is performed in the scene, resulting in mixed illumination information and cause the difficulty to undertake illumination editing tasks. Neural Reflectance Field~\cite{Relighting:ReflectanceField} is the first method to separate illumination information in the implicit field. It assigns each sampling point on the ray a ``visibility'' to reach the light source, and then uses this visibility to control the color expression of the spontaneous spot to achieve relighting task. PhySG~\cite{Relighting:physg} and NeRD~\cite{Relighting:nerd} methods ignore the self-occlusion of the single target and represent the ambient light and BRDF as the mixture of spherical Gaussian functions. NeRV~\cite{Relighting:nerv} establishes the implicit field on the 3D spatial coordinates and the 2D incidence directions(the 2 angles), and produces a variable reflecting the visibility of corresponding light source. NeRFactor~\cite{Relighting:nerfactor} starts with a pre-trained NeRF model, reduces the geometry to a single surface and outputs the visibility toward light source and normal factor against the surface. Considering the unknown illuminating condition in our work, we adopt the method based on the depth~\cite{NeRFVariant:spinnerf} to calculate the ``depth map'' from the light source direction for flexible shadow editing results, so as to supervise the shading layer. A reasonable shadow from direct light can be obtained after editing by artificially set illuminating conditions.


\section{OSI-aware Scene Reconstruction and Editing System}
\subsection{Overview}
\begin{figure*}[htbp]
	\begin{center}
		\includegraphics[width=\textwidth]{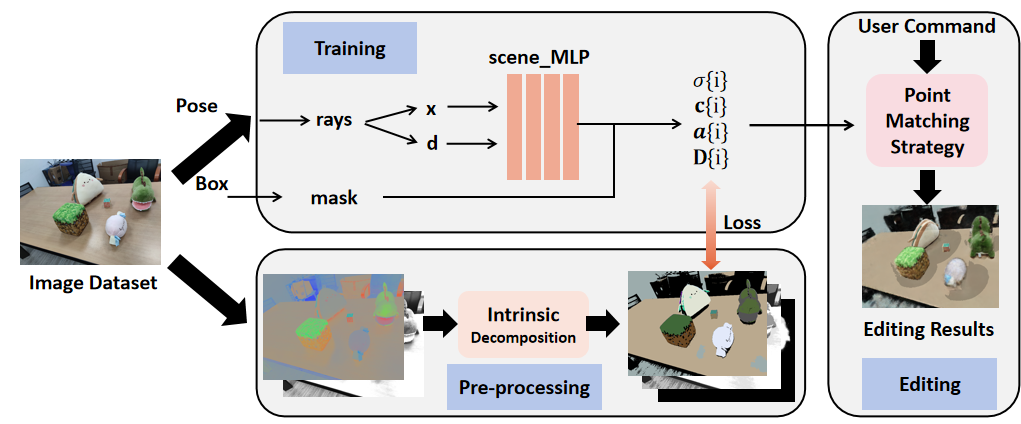}
	\end{center}
	\caption{An overview of our proposed OSI-aware system. For the input image $\mathcal{I}$, we first perform intrinsic decomposition, separating albedo images and shading images as supervision in training stage. Then we parameterize the neural implicit field. It maps position $x$ and direction $d$ to properties including $\sigma$, color $\textbf{c}$, albedo $\textbf{a}$ and depth $\textbf{D}$ at sampling points on the rays. We also use 2D and 3D mask on the rays to segment the editable objects $i$ from background. In the editing stage, the system are guided by user's commands to recognize and draw shadow areas, and produces shadow results that can be freely edited in direction, intensity and range.}
	\label{fig:overall flow}
\end{figure*}

As shown in Fig. (\ref{fig:overall flow}). Our method consists of three parts. At the data preprocessing stage, we conduct data preprocessing through a prior guided intrinsic decomposition method, obtaining the albedo, shading and residual supervision required in neural network learning. In the training stage, we control the ray sampling through 2D mask and 3D mask to guide the network to produce the required editable objects and background envrionment information by implicit field. In the final editing stage, we fill the ``holes'' in the background with a depth-guided approach and guide the system to freely produce editable shadow results through the point matching strategy.

\subsection{Prior Guided Image Intrinsic Decomposition} \label{sec:valid_analysis}

According to Retinex theory~\cite{IntrinsicDecomposition:Retinex}, an image contains many different layers. Subsequent researchers chose the two most significant layers, the albedo layer and the shading layer, to simplify the problem. In most of the related works~\cite{IntrinsicDecomposition:2D1,
IntrinsicDecomposition:2D2,
IntrinsicDecomposition:2D3,
IntrinsicDecomposition:2D4}, the albedo layer often depicts the slow gradient changes in the image, which often occurs in the neighborhood of the same object's surface and stays invariable despite illuminating condition change in the scene. The shading layer often depicts the sharp gradient changes in the image, which occurs in shadow edge or texture of targets and indicates the targets' illuminating conditions. We follow the similar path and add a third residual layer in our work, just like Ye \etal~\cite{NeRFVariant:intrinsicnerf}. The albedo part is assumed as the scene under no direct or indirect illumination while the shading part contains all the illuminating information. All the texture details are demonstrated in the third residual part, which contains all the ``small swings'' superimposed on the smooth and stable product of albedo and shading. The decomposition formula is shown as follows:

\begin{equation} \label{equ:intrinsic_decomposition_image}
	I = A \odot S + Re,
\end{equation}

The $\odot$ is element-wise product. The decomposition process is ill-posed, because we hope to obtain the color of the scene under no direct or indirect illuminating conditions through it. The color is missing in the datasets, so there may be many different reasonable outputs. Our goal is separating the shadows and specular factors from the images as much as possible, so we follow a grayscale shadow hypothesis and a brightness prior.

The grayscale shadow hypothesis holds that the shading layer of images are single channel. The shading layer's nature is a illumination factor superimposed on the albedo layer, indicating the illumination effects of the pixel corresponding to the target. Therefore we set the shading output of the neural network to single channel.

The brightness prior believes that the shadow and the specular factor is represented in the image as a variation of the image brightness. By normalizing the brightness of image pixels, most of the shadow effects can be removed to ensure the recovery of albedo colors.

\begin{figure}[htbp]
	\begin{center}
		\includegraphics[width=\linewidth]{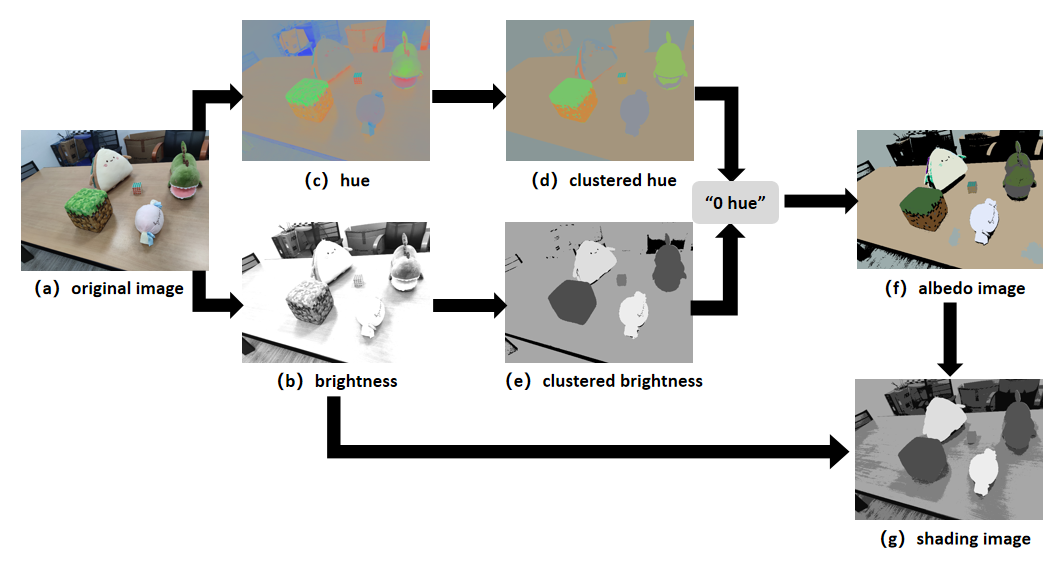}
	\end{center}
	\caption{The intrinsic decomposition results through grayscale shadow hypothesis and brightness prior are shown in the figures. Fig. (a) is the original image, Fig. (b) is the single-channel brightness, Fig. (c) is the hue image, Fig. (d) and Fig. (e) are the unsupervised clustering normalized hue and brightness, Fig. (f) is the pixel-wise product of Fig. (d) and Fig. (e), representing the albedo layer of the image. Fig. (g) is the shading layer of the image.}
	\label{fig:intrinsic_decomposition_result}
\end{figure}

\begin{figure}[htbp]
	\begin{center}
		\includegraphics[width=\linewidth]{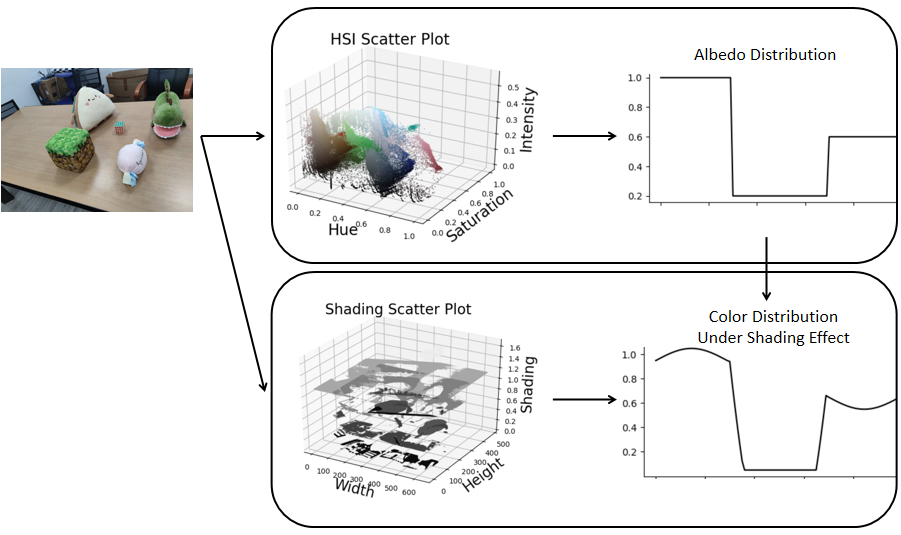}
	\end{center}
	\caption{The numerical relationships among the albedo, shading and residual layers is shown in the Fig. (\ref{fig:intrinsic_decomposition_result}). The image on the left is the original image. For- a given scene we extract its color distribution, which is complex and diverse. After the intrinsic decomposition, the albedo color is clustered according to the hue transformation. Our goal is to cluster the image into several color-consistent blocks like the plot on the right. Within each hue cluster, there are shadings representing occlusions. Our shading cluster is to get the ``bright and dark'' effect within the same color block according to the color blocks obtained by albedo. The shading superimposing on albedo is shown on the right, we can see the difference where the original color is consistent, but this difference does not lead to color mixing. The general color distribution remains. The residual is indicated as a small fluctuation on it, representing details and texture of the image.}
	\label{fig:intrinsic_decomposition_numeral_analysis}
\end{figure}

The intrinsic decomposition results are shown in the Fig. (\ref{fig:intrinsic_decomposition_result}). Fig. (\ref{fig:intrinsic_decomposition_numeral_analysis}) shows the numerical relationships among the image layers obtained after the intrinsic decomposition. \textbf{Fig. (a)} is the original image. \textbf{Fig. (b)} is the single-channel brightness grayscale map generated according to the grayscale shadow hypothesis, which shows the remarkable drop in value within shadow areas of the image. However, there are black objects in the scene whose grayscale values are also close to 0, making it insufficient to guide the intrinsic decomposition process alone. Therefore, we adopt a form of two-way decomposition: similar to the idea of HSI transformation, we transit the rgb color into another color space, the right decomposition form will be found in the process. In our work, we divide the image into two parts of hue and brightness, and performed unsupervised clustering on them respectively. \textbf{Fig. (b)} represents the brightness of the image and \textbf{Fig. (c)} represents the hue, that is, the original \textbf{Fig. (a)} divided by brightness \textbf{Fig. (b)}. We found that this division suppress the brightness change of the scene due to shadowing and eliminate the interference of black objects in the decomposition process.

On the basis of \textbf{Fig. (b)} and \textbf{Fig. (c)}, unsupervised clustering of mean-shift algorithm is carried out to ensure the consistency of overall brightness and hue, the results are shown in \textbf{Fig. (d)} and \textbf{Fig. (e)}. \textbf{Fig. (f)} is the pixel-wise product of \textbf{Fig. (d)} and \textbf{Fig. (e)}, which is the albedo image we acquire from the decomposition process. For black objects, we found in the experiment that due to its lower brightness the various external factors, the captured images usually can not obtain their hue well, so we introduce a ``0 hue'' with rgb values of 0 in the clustering process, and keep the black objects' color unchanged. In the process of brightness clustering, the computer will automatically fine-tune the number of clusters and get the clustering results including shadows and the clustering results without shadows. The former becomes shading layer of the intrinsic decomposition output, as shown in \textbf{Fig. (g)}. Finally, according to Equ. (\ref{equ:intrinsic_decomposition_image}), residual image will be obtained by the difference between original \textbf{Fig. (a)} and the pixel-wise product of albedo image \textbf{Fig. (f)} and shading image \textbf{Fig. (g)}

\subsection{Mask Guided OSI-aware Implicit Field Reconstruction} \label{sec:network_analysis}

Given a set of complex indoor scene images with unknown illuminating conditions, our goal is to separate the basic attributes of the scene, such as the albedo, shading and texture(residual), and achieve object-level reconstruction and editing of the background and editable targets in the scene. Fig. (\ref{fig:overall flow}) depicts the outline framework of our proposed method. We perform two tasks: (1) We guide the rays to reconstruct the editable implicit neural field of the scene by 3D box supervision, respectively rebuild the volume density \textbf{$\sigma_i$}, color \textbf{$c_i$} and albedo \textbf{$a_i$}. (2) We use 2D image supervision $I_{instance}$ and the recovered volume density information \textbf{$\sigma_i$} to generate a ``shadow region'' along with the rays. Experiments show that our method can be applied to unknown scenes under any illuminating conditions and produce reasonable interaction between objects and scene environment.

\subsubsection{Implicit Representation.}

\begin{figure*}[htbp]
	\begin{center}
		\includegraphics[width=\textwidth]{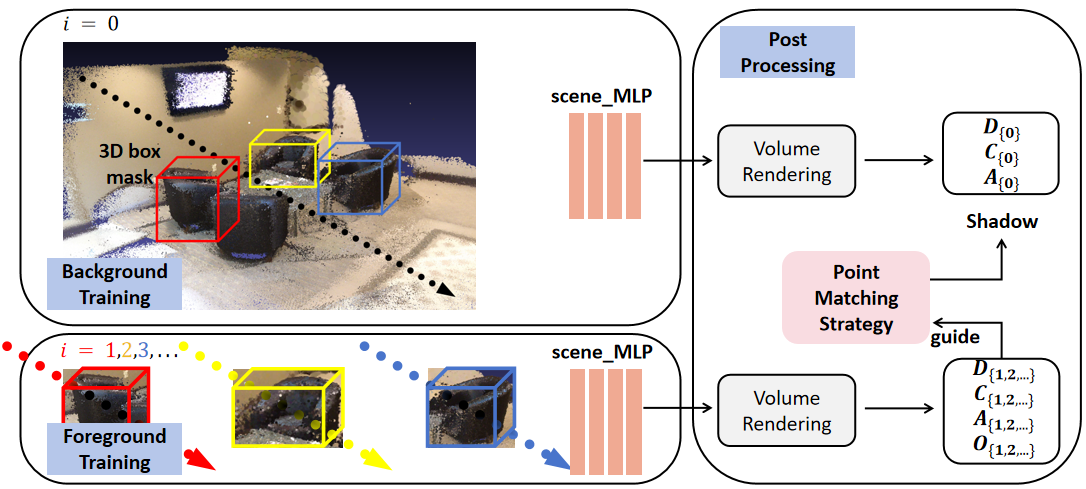}
	\end{center}
	\caption{The training stage using our proposed two-stream neural network. \textbf{Background training:} the embedded spatial point vectors $\gamma(x)$ and orientation vectors $\gamma(d)$ are sent to the network. The network provides volume density $\sigma$, true image color $C$ and albedo $A$ as outputs. \textbf{Foreground training:} In this part we use 3D box to split the rays and set all the $\sigma$ corresponding to points outside the box to 0. \textbf{Post processing:} After volume rendering through ray casting, the system output the depth information $D$ and use it to guide the shadow network producing the shadow information.}
	\label{fig:scene netwrok}
\end{figure*}

We propose an implicit field representation that integrates both albedo and rgb color layer, and foreground-background segmentation into the neural network, as shown in Fig. (\ref{fig:scene netwrok}). Our NeRF variant splits the scene into object targets and background by 3D box supervision. In the data preprocessing stage, we calculate the near and far distance of the rays passing through 3D box while generating rays. When querying the implicit field, we first sample the points along the rays with the same near and far distance, and then segment the queried foreground and background volume density field by the ranges we have obtained from 3D box. The scene network model is denoted as $\mathcal{F}_{scene}(\gamma(\textbf{x}), \gamma(\textbf{d}))$. We encode the spatial coordinates $\gamma(\textbf{x})$ and the direction coordinates $\gamma(\textbf{d})$ to obtain high-dimensional embedded representation.

\begin{equation} \label{equ:scene_branch_1}
\sigma_{i}(\textbf{r}) = \textbf{F1}_{scene}(\gamma(x))
\end{equation}

\begin{equation} \label{equ:scene_branch_2}
c_{i}(\textbf{r}), a_{i}(\textbf{r}) = \textbf{F2}_{scene}(\gamma(x), \gamma(d))
\end{equation}

\textit{Volume Rendering.} After obtaining the 3D attributes of the sampling points \textbf{$p$} on the rays \textbf{$r$}, the corresponding opacity \textbf{$o$}, color \textbf{$c$}, albedo \textbf{$a$}, depth \textbf{$d$} and shading \textbf{$s$} can be obtained by color integration~\cite{} of volume density \textbf{$\sigma$}. The \textbf{$\sigma$} represents the differential probability of a ray of light passing through space and terminating at an infinitesimal point \textbf{$p$}. The rays have boundaries, expressed by \textbf{$t_n$} and \textbf{$t_f$}.

\begin{equation} \label{equ:integration_delta}
\delta_k = t_{k + 1} - t_k, 
\alpha(\delta_k) = 1 - e^{-\delta_k\sigma(t_k\delta_k)}
\end{equation}

\begin{equation} \label{equ:integration_T}
T(\delta_{k}) = \prod \limits_{k=0}^{k-1} (1 - \alpha(\delta_{k}))
\end{equation}

Where $k$ is the sequence number of sampling points from near to far along the ray, and $\delta$ represents the distance between two adjacent sampling points. $\alpha$ describes the probability that the light will terminate at a specific sampling point. The meaning of $T$ is the probability that the light will pass through all the previous points and continue through that point on the ray. The probability that light goes from $t_1$ to $t_k$ is the same as it emitted backwards, so we can not only perform ray casting for novel views, but also for novel illuminating conditions.

\subsubsection{Loss function}

Since we split the implicit field through 3D box, the resulting loss function needs to be applied to both foreground and background reconstruction. In the foreground reconstruction, we first use the 3D box mask on the rays to separate editable objects from the scene, as shown in the equation Equ. (\ref{equ:separate_frontback}). Then set the color $\textbf{C}$ as the training target, as shown in the equations Equ. (\ref{equ:integration_C_f}) and Equ. (\ref{equ:loss_fore}).

\begin{equation} \label{equ:separate_frontback}
M(n) = 1 - \frac{1}{1 + e^{-\lambda_{1}(t_k - t_n)}}, 
M(f) = 1 - \frac{1}{1 + e^{-\lambda_{1}(t_f - t_k)}}
\end{equation}

We set $\lambda_{1}$ as 10.0. When generating the rays, we have ensured that the nears of the rays exist only at the contact surface between 3D box and the background scene. Considering the mutual occlusion between the targets, the near distance can only be set to the contact surface closest to the view point $t_n$. Therefore, it is safe to use this prior to optimize the spatial distribution of the volume density field. So when we in the foreground reconstruction stage, we inhibit the expression of the volume density during the volume rendering process.

\begin{equation} \label{equ:integration_C_f}
C(\textbf{r}) = \sum \limits_{k=1}^K (1 - M(f)) (T(\delta_k)\alpha(\delta_k)c(t_k))
\end{equation}

\begin{equation} \label{equ:integration_O}
O(\textbf{r}) = \sum \limits_{k=1}^K (1 - M(f)) (T(\delta_k)\alpha(\delta_k))
\end{equation}

\begin{equation} \label{equ:integration_O_n}
O_n(\textbf{r}) = \sum \limits_{k=1}^K M(n) (T(\delta_k)\alpha(\delta_k))
\end{equation}

\begin{equation} \label{equ:mse}
MSE(x, y) = \frac{1}{N} \sum \limits_{i=1}^N (x - y)^2
\end{equation}

\begin{equation} \label{equ:loss_fore}
\mathcal{L}_{fore} = MSE(C(\textbf{r}), \textbf{C}) + MSE(O(\textbf{r}), Mask) + \lambda_{2}MSE(O_n(\textbf{r}), 0)
\end{equation}

where $N$ represents the batch size, $Mask$ represents the 2D instance mask we have, We set $\lambda_{2}$ as 0.001 to match the magnitudes. By these we utilize the 3D box of the foreground and perform ray casting and color integration from the near points $t_n$ to the far points $t_f$, as well as from the camera center $t_1$ to the near points $t_n$. The latter part was suppressed on opacity $O(\textbf{r})$ because of the synchronously produced prior condition we have when generating rays, that is, there is no object occlusion from camera center to the near points otherwise the near points will be set on the occluded part. Therefore we perform additional supervision on the foreground reconstruction, as shown by Equ. (\ref{equ:integration_O_n}), improving the geometric accuracy.

The color and geometric reconstructions of the background are similar to the way we performed in the foreground reconstruction. However this time we don't use 3D box masks to control the integration process along the rays. The color and albedo results are shown by Equ. (\ref{equ:integration_C}) and Equ. (\ref{equ:integration_A}).

\begin{equation} \label{equ:integration_C}
C(\textbf{r}) = \sum \limits_{k=1}^K (T(\delta_k)\alpha(\delta_k)c(t_k))
\end{equation}

\begin{equation} \label{equ:integration_A}
A(\textbf{r}) = \sum \limits_{k=1}^K (T(\delta_k)\alpha(\delta_k)a(t_k))
\end{equation}

\subsection{Depth Guided Post-processing and Point Matching Scene editing Strategy.} \label{sec:test_analysis}

In the background reconstruction, we not only need to improve the color and geometric distribution, but also need to use the depth obtained by the removing the editable objects' geometric distribution to guide the inpainting of image ``holes''.

\begin{equation} \label{equ:integration_D}
D(\textbf{r}) = \sum \limits_{k=t_f}^{K} (T(\delta_k)\alpha(\delta_k)t_k)
\end{equation}

\begin{figure}[htbp]
	\begin{center}
		\includegraphics[width=0.7\linewidth]{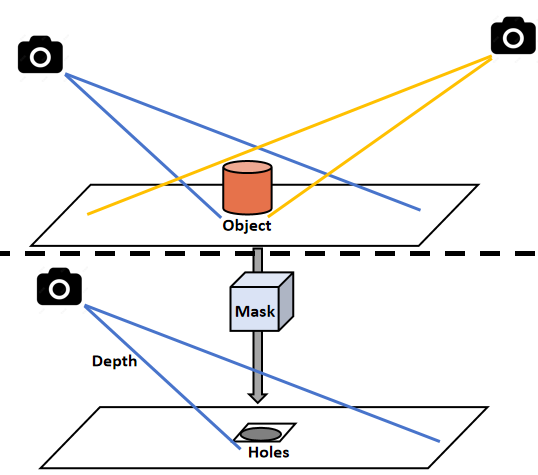}
	\end{center}
	\caption{The scene inpainting process. \textbf{Upper part} shows the reason why editable objects leave ``holes'' in the background: The editable objects are fixed in the 3D scene, there will always be an unsupervised part of the contact surface between the object and the background during training. \textbf{Lower part} shows how we use the depth map to guide color filling when we mask the editable object from the scene.}
	\label{fig:scene inpainting.}
\end{figure}

The editable objects did not move, as shown by Fig. (\ref{fig:scene inpainting.}). There are always unsupervised regions like the contact surfaces between the objects and the background, which exists in none of the datasets. Therefore, we have to perform an associative method to fill the ``holes''. Inspired by ~\cite{NeRFVariant:spinnerf}, the association basis is depth, which we obtained from the equation Equ. (\ref{equ:integration_D}). We calculate the depth and measure the differences between the pixels and the corresponding pixel parts of the ``holes'' on the whole depth map, and assigns weights $\textbf{w}_{depth}$ on every pixel. The smaller the depth differences the larger the weights $\textbf{w}_{depth}$. The weights corresponding to the pixels of the ``holes'' are set to 0. We normalize and multiply the calculated weights with the object-removed albedo image, then fill the color of ``holes'' with the weighted value sum.

\begin{equation} \label{equ:depth_weight}
\textbf{w}_{depth} = 1 - e^{\lambda_3[\lvert D(\textbf{r}) - D(\textbf{r}_\textbf{H}) \rvert]}
\end{equation}

\begin{equation} \label{equ:depth_weight_norm}
\textbf{w}_{depth} = \frac{\textbf{w}_{depth}}{\sum \limits_{i=1}^N \textbf{w}_{depth}}
\end{equation}

We set $\lambda_3$ to -10.0, N represents the batch size. \textbf{H} is a pixel set, which contains all the ``hole'' pixels in the albedo result. After that we use the MSE loss as shown by Equ. (\ref{equ:mse}) to supervise the $\textbf{F}_{scene}$ model generating reasonable albedo results for the ``hole'' pixels.

\begin{figure}[htbp]
	\begin{center}
		\includegraphics[width=\linewidth]{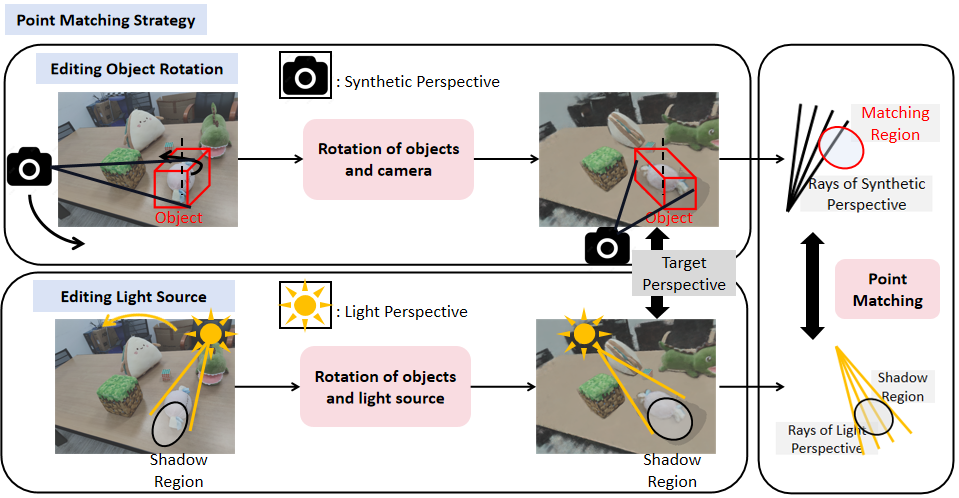}
	\end{center}
	\caption{\textbf{Point matching strategy:} In the same way that we emit rays from the camera and calculate the integral to finally generate the \textbf{synthetic perspective}, we also set up the artificial light source and emit the rays. By calculating whether the light passes through the object we can find the ``occluded'' points, we name the rays \textbf{light perspective}. In the editing task, the rotation or movement of editable objects will affect their relative position in the light perspective. We change the light perspective by rotating or moving the light source to achieve a relative position, that the object is at, in the light perspective after editing. Simply put, move the light perspective and the synthetic perspective into the same original set. Thus, we obtain the ``occluded points'' of the light perspective, and then match them with the synthetic perspective to find their corresponding shadow points, so that we can get the corresponding shadow area in the rendering image.}
	\label{fig:shadow rendering.}
\end{figure}

In our post-processing module, there is a shadow rendering part in addition to scene inpainting. We use the point matching strategy to guide the shadow rendering. As shown in Fig. (\ref{fig:shadow rendering.}), in the editing phase, we usually use ray casting algorithm to index the implicit field to present synthetic views. Our shadow rendering method is inspired by it. First, we artificially set up an artificial point light source $\textbf{L}_{a}$ in the scene and emit rays $\textbf{R}_{l}$ through the point light source. Such rays do not necessarily make up the image, so we create them using the points of the editing target. We can obtain the points by 3D boxes on the synthetic sampled points $M(n)\textbf{P}_{view}$. After the rays are emitted, we sample points on the ray, calculate the transmittance $T$ against the editable object by volume rendering, and find a rough ``occlusion'' section $M_{occ}$ along the rays. Finally, we use a simple point matching method to find the points $P_{match}$ from the sampled rays of the scene environment in the test stage, and assign them a value representing the intensity of the shadow (or occlusion). We artificially control the threshold of matching and the intensity of shadows through parameter $\beta$ and $\gamma$ to achieve free control of shadow intensity and shadow range.

\begin{equation} \label{equ:rays_emmit}
\textbf{R}_{l} = M(n)P_{view} - \textbf{L}_{a}
\end{equation}

where we choose ray $\textbf{r}$ in the rays set $\textbf{R}_{l}$. Then we sample points along the ray by index $k={0,1,2,...}$. As shown in Equ. (\ref{equ:integration_T}), the transmittance $T(k)$ is obtained on every sample points $P_k(\textbf{r})$.

\begin{equation} \label{equ:occlusion}
M_{occ}(\textbf{r}) = \{k | T(k) < \gamma\}
\end{equation}

$\gamma$ is a threshold between 0 and 1 that will control how many of the sample points are considered to be obscured by objects. The smaller the $\gamma$, the more stringent the determination. Ablation experiments have shown that it is set to be between 0.1 and 0.3, we can obtain best occlusion result.

\begin{equation} \label{equ:point_matching}
P_{match}(\textbf{r}) = \{p | min(\sqrt{||{M_{occ}(\textbf{r})P_k(\textbf{r}) - p}||}) < \beta \}
\end{equation}

where $\beta$ represents the matching threshold of points between the synthetic view and the artificial light rays view. It is generally set to around 0.05 in the Toydesk Dataset (depending on the scale of the whole scene). This threshold controls the range of the shadow. If the threshold is too large, some un-occluded areas will also produce shadows. If the threshold is too small, the shadow range may be defective. In the editing stage, it is possible to edit the position and direction of the objects. At this time we only need to transform the direction of the rays $\textbf{R}_{l}$ to $\textbf{R}_{l}^{edit}$, and finally we apply the occlusion mask $M_{occ}^{edit}$ back to $\textbf{R}_{l}$ for the next point matching process.

In the scene editing stage, for the rotation and shift editing tasks, we rotate or move the camera rays around the editable targets, and obtain the edit targets after rotation or shift by integration on the rays. For the addition of a new target in the scene, we transform the spatial coordinate system of the scene to the spatial coordinate system of the new target, obtaining the rendering result by adjusting the sampling strategy. Specifically, we increase the density of point sampling within the selected 3D box. 

For the interaction between objects and scene environment, like shadows, we emit rays from the setting light source, then calculate the transmittance $T$ of the ray on the editable object to find the occluded points. Finally, the shadow is drawn by corresponding points in the scene environment obtained by point matching.


\section{Experiments} \label{sec:experiments}

\subsection{Experimental Overview}
We first introduce 3 datasets: Toydesk, Scannet and Replica, which are all indoor datasets. The comparison among datasets are shown in Tab. (\ref{tab:brief_of_all_dataset}). Toydesk is a real scene dataset while the other 2 are synthetic scene datasets. 

Next, we study the separation capability of our proposed system of interaction between objects and scene environment from both qualitative and quantitative perspectives. Qualitatively, we compare the results before and after the image separation, including scene inpainting and specular factor removal. Quantitatively, we compare the differences of corresponding feature points after separating images from different perspectives to verify the effectiveness and stability of the intrinsic decomposition method we apply. 

After that, we show applications of our model on scene editing tasks on object rotation and light source editing. Then we calculate the rendering quality of the model compared to other methods.

Finally we perform rigorous ablation study to analyze the different parts of our system. We compare the ability of other intrinsic decomposition methods to extract interaction between objects and scene environment information. The results of scene editing with and without scene inpainting methods are also studied, both quantitatively and qualitatively.

\begin{table}[htbp]
	\begin{center}
		\caption{Brief comparison among different datasets.}
		\label{tab:brief_of_all_dataset}
		\setlength{\tabcolsep}{2mm}
		\begin{tabular}{lcccc}
			\toprule[1.3pt]
			Dataset & Source & Image & 3D box label & Semantic label \\
			\hline
			\specialrule{0em}{1pt}{1pt}
			$\mathcal{D}_{toydesk2}$ & Yang \etal~\cite{NeRFVariant:objectnerf} & 151 & $\checkmark$ & $\checkmark$ \\
			$\mathcal{D}_{scannet113}$ & Dai \etal~\cite{dai2017bundlefusion} & 733 & $ \checkmark$ & $\checkmark$ \\
			$\mathcal{D}_{replica}$ & Julin \etal~\cite{replica19arxiv} & 900 & $\times$ & $\times$ \\
			\bottomrule[1.3pt]
		\end{tabular} 
	\end{center}
\end{table}

\subsection{Intrinsic Decomposition Dataset Generation}  \label{sec:training_data_gen}

\begin{figure}[htbp]
	\begin{center}
		\includegraphics[width=\linewidth]{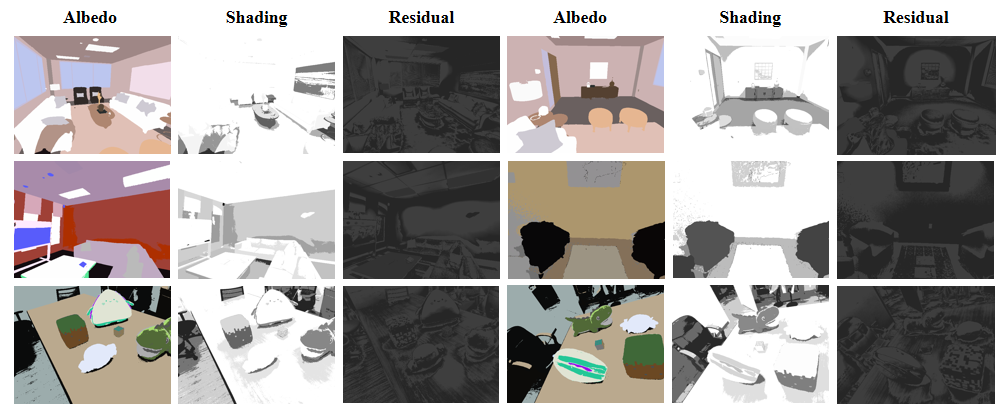}
	\end{center}
	\caption{Illustration of the generated intrinsic decomposition dataset. There are several editable objects and a non-editable indoor scene. We mask the targets and perform intrinsic decomposition on background and foreground respectively. Here we show six sets of images obtained through intrinsic decomposition, including the albedo image, the shading image and the residual image.}
	\label{fig:dataset_visual}
\end{figure}

Through the six intrinsic decomposition results in Fig. (\ref{fig:dataset_visual}), we can assume that in the Albedo part, we only keep relatively reasonable 3-channel color information, in the Shading part, we keep all the shadow information in a single-channel grayscale map. The Residual part is more complex and detailed, equivalent to a small scale fluctuation superimposed on the image. Due to its small magnitudes we increased the grayscale value by 50 (255 is the full grayscale value) to illustrate the details of Residual image.

To train our model, a training set with 2D tags and 3D box labels is necessary. The real deatset Toydesk has these two traits, and for the synthetic datasets Scannet and Replica, we browse the indoor scene and generate labels through the habitat simulator. In order to obtain the separated interaction between objects and scene environment information, we also need to perform intrinsic decomposition method on the 2D image the obtain its albedo, shading and residual images. We accomplish this automatically with python code.

\subsection{Experimental Setups}
We train the model on the synthetic dataset and then complete the editing task on the corresponding dataset to evaluate the performance. Specifically, we use the data generated by the intrinsic decomposition method as supervision, directing the background branch to produce a special implicit field. In the editing stage, we query this implicit field through the sampling points to obtain the corresponding properties, which are then projected into the 2D image plane through the volume rendering. All of our editing tasks are based on this process. In order to speed up the training process, we reduce the size of the input image to half, and then obtain the training data through random selection  by pytorch-lightning. To evaluate the rendering quality, we use the PSNR, SSIM~\cite{wang2004SSIM} and LPIPS~\cite{zhang2018LPIPS} metrics. 

In this paper, our experiments consist of three parts. First, the scene editing task. We designed and carried out tasks such as object movement, rotation and light field editing through our system, and implemented scene inpainting and shadow rendering through the post-processing module, so as to produce reasonable shadow effects while editing. In the second part, we compared our work with other NERF-based image rendering methods and confirmed that our method provided competitive results in image rendering quality. In the third part, we conducted ablation experiments to verify the effectiveness of each module we designed.

\subsection{Scene editing tasks.}
\begin{figure}[htbp]
	\begin{center}
		\includegraphics[width=\linewidth]{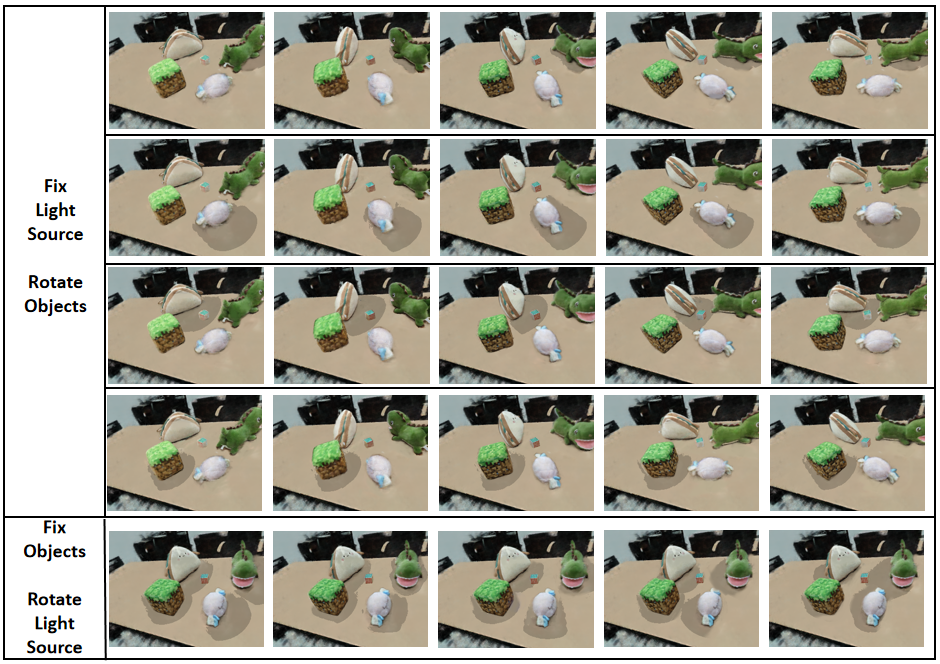}
	\end{center}
	\caption{Illustration of object rotation and corresponding shadow changes for the Toydesk dataset. For each object we keep the position and direction of the point light source unchanged, rotate the object, and automatically draw the corresponding shadow. For the light field, we fixed the object, and plotted the corresponding shadow changes by changing the position and orientation of the artificial point light source.The editing results show that compared with other editing methods, like~\cite{NeRFVariant:objectnerf}, our method can draw shadows that rotate with the rotation of the editable object, and can also change along with the position and orientation of the artificial light source and the shape of the editable object.}
	\label{fig:edit_tasks}
\end{figure}

In this m we use our model to complete various high-level editing tasks: object movement, object rotation, object copy, new object addition and light field editing. Fig. (\ref{fig:edit_tasks}) shows the results of two special tasks: object rotation and light field editing.

\

Our editing method mainly lies in the selection of ray sampling points. In the task of object movement and rotation, we rotate the rays of the synthetic view around the center of the object to be edited, and use 3D mask to guide the computer to find the position of the editable object along the ray (one-dimensional coordinates). Then we carry out color integration according to the order of one-dimensional coordinates. Similarly, we find the one-dimensional coordinates of the shadow region on the ray by matching the ray points, and then draw the shadow in the synthetic view. In the task of adding a new object, we keep the rays unchanged and transform the original rays into the corresponding coordinate system of another implicit field (the object to be added) by means of a coordinate transformation. We perform the corresponding transformation through a w2w matrix, which is calculated by the c2w matrix corresponding to two different scenarios.

\subsection{Comparison with the State-of-the-Art Methods.}\label{sec:cmp_sota}
\begin{figure}[htbp]
	\begin{center}
		\includegraphics[width=3in]{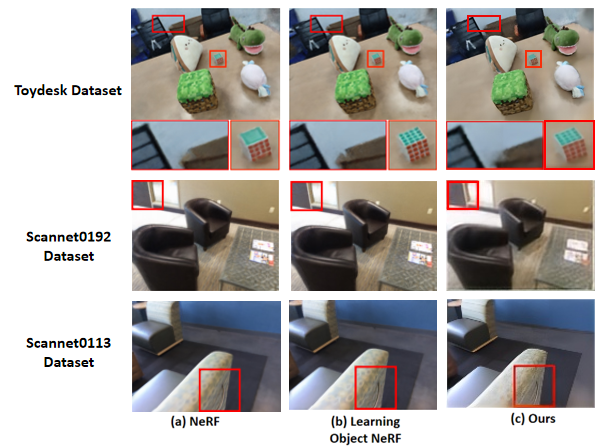}
	\end{center}
	\caption{We compare scene rendering quality with NeRF and Learning Object NeRF on the ToyDesk and Scannet dataset. Our method maintains relatively good rendering quality. The contents within red boxes show the details of the rendering result, which allows the readers to compare the rendering fineness produced by different methods. For example, in the Toydesk dataset (the first row in the figure), the NeRF method shows the least fineness, Learning Object NeRF has the best results, and our method lays in between.}
	\label{fig:sota}
\end{figure}

In this subsection, we compare our method against state-of-the-art methods by rendering quality (NPCR~\cite{NeRFVariant:NPCR}, NeRF~\cite{NeRFVariant:NeRF}, Learning Obj NeRF~\cite{NeRFVariant:objectnerf}) on the background branch and foreground branch respectively. The qualitative and quantitative comparison results are shown in the Fig. (\ref{fig:sota}) and Tab. (\ref{tab:rendering_quality_comparision}). Fig. (\ref{fig:sota}) qualitatively shows the render quality of our method in detail compared to NeRF and Learning Object NeRF methods, while Tab. (\ref{tab:rendering_quality_comparision}) shows the overall score of our method in PSNR, SSIM, and LPIPS metrics compared to other methods.

\begin{table}[htbp]
	\begin{center}
		\caption{Rendering Quality.}
		\label{tab:rendering_quality_comparision}
		\setlength{\tabcolsep}{2mm}
		\begin{tabular}{lcccccc}
			\toprule[1.3pt]
			\multicolumn{ 1}{l}{} & \multicolumn{ 3}{c}{Toydesk} & \multicolumn{ 3}{c}{Scannet} \\
			Config & PSNR$\uparrow$ & SSIM$\uparrow$ & LPIPS$\downarrow$ & PSNR$\uparrow$ & SSIM$\uparrow$ & LPIPS$\downarrow$ \\
			\hline
			\specialrule{0em}{1pt}{1pt}
			NPCR~\cite{NeRFVariant:NPCR} & - & - & - & 25.177 & 0.754 & 0.225 \\
			NeRF~\cite{NeRFVariant:NeRF} & 15.453 & 0.586 & 0.537 & 28.927 & 0.815 & 0.249 \\
			Learning Object NeRF~\cite{NeRFVariant:objectnerf}(Scene Branch) & 15.607 & 0.532 & 0.522 & 29.005 & 0.815 & 0.243 \\
			Ours & \textbf{23.049} & \textbf{0.932} & \textbf{0.030} & \textbf{30.615} & \textbf{0.936} & \textbf{0.067} \\
			\bottomrule[1.3pt]
		\end{tabular} 
	\end{center}
\end{table}

\subsection{Ablation Study} \label{sec:ablation_study}

In this subsection, we evaluate the effectiveness of 1) Shadow rendering by point matching strategy, 2) our intrinsic decomposition method, and 3) analyze the scene inpainting.

\textbf{Analysis of shadow rendering.}
In order to evaluate the effectiveness of our shadow rendering method, we conducted an ablation experiment based on our shadow rendering process. Our shadow rendering method is inspired by the idea of ray casting~\cite{NeRFTechnique:ray}. We verifies whether the the sampling points along the rays are occluded by editable objects by the transmittance $T$, given by Equ. (\ref{equ:integration_T}). We match the points considered to be occluded with the ray sampling points from image rendering view to find the ``shadow area'' in the background scene. We set a threshold $\beta$ for the transmittance $T$ to control the range of ``blocked'' region, and a threshold $\gamma$ for point matching to control the shadow area at the ray sampling point from the image rendering view. In the editing stage, we manually control the direction and position of the point light source to generate shadows of editable objects, as well as the area and scope of the shadows.

\begin{figure}[htbp]
	\begin{center}
		\includegraphics[width=\linewidth]{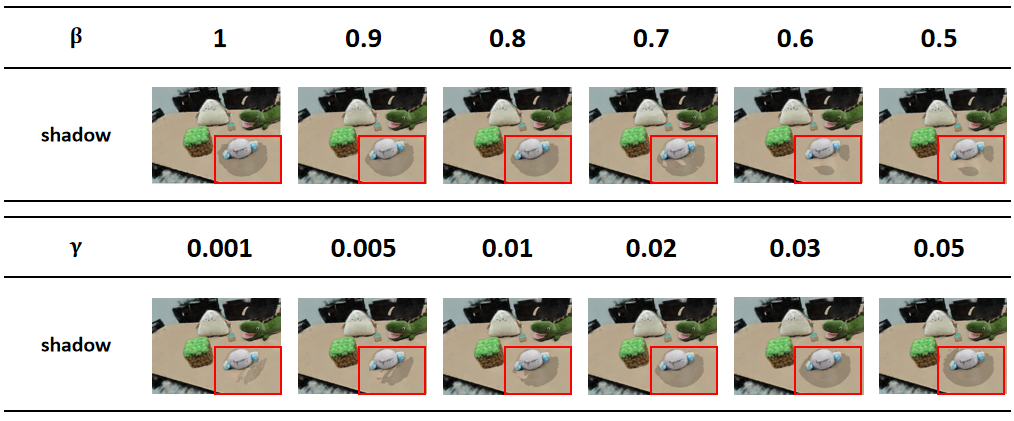}
	\end{center}
	\caption{Illustration of the effect of the two thresholds $\beta$ and $\gamma$ we set on the shadow rendering process. To visualize the effect of our shadow rendering method on a complex indoor scene, we used the Toydesk dataset~\cite{NeRFVariant:objectnerf} and selected one of the editable objects to directly show the qualitative result of shadow rendering under different threshold Settings. $\beta$ is the threshold of transmittance $T$, determining which sample points of light emitting rays fall into the ``blocked region''. While $\beta$ is setting too low, it will result in a fragmented distribution of shaded area. $\gamma$ is the Euclidian distance threshold for point matching, determining whether the synthetic perspective has a similar distribution of sample points to the occluded points from light perspective. While $\gamma$ is setting too low or too high, it will affect the judgment process, causing the shaded area to corrode or dilate.}
	\label{fig:shadow_parameter_compare}
\end{figure}

Let $\beta$ represent the threshold of transmittance $T$. In the ray casting algorithm~\cite{NeRFTechnique:ray}, transmittance is considered to be the probability that the light will continue to pass through after reaching a certain place from the ray center. When the transmittance is low, the place is considered occupied. At higher transmittance, the place is considered transparent or empty. In our shadow rendering method, we are inspired by the ray casting algorithm. So we emit rays from an artificial light source and compute the transmittance along the rays. In Fig. (\ref{fig:shadow_parameter_compare}), we can see that when the threshold $\beta$ is set between 0.9 and 0.8, the shadow we draw will have a relatively better result. When $\beta$ is lower than 0.8, some of the occluded areas are considered 
``not occluded'', and when $\beta$ is higher than 0.9, some of the previously unobstructed areas are considered ``occluded''.

Let $\gamma$ represent the threshold we set for the point matching. In our shadow rendering process, the rays we emit from the artificial light source do not necessarily follow the way the image is rendered (in other words, each rays needs to pass through each pixel in the image plane). More simply, we identify a rough ``shadow region'' in the 3D space, and then find the sampling points that fall in this region in the rays that make up an image. Since we determine this region by sampling points on the rays emitted from the light source, we need to find the point meeting the requirement by point matching method.

We use the Euclidean distance from point to point to find the sample points that meet the requirement, and control the matching degree by threshold $\gamma$. As we can see in Fig. (\ref{fig:shadow_parameter_compare}), when $\gamma$ threshold is low, some points in the shadow area will not match, while when the $\gamma$ is too high, some points outside the shadow area will fall into the region. Ablation experiments show that this threshold is related to the scale of the scene. In Toydesk's scenario, the most appropriate $\gamma$ is considered to be around 0.02.

\textbf{Intrinsic decomposition.}
To verify the effectiveness of the intrinsic decomposition method we used, we compared the albedo images obtained by Intrinsic NeRF~\cite{NeRFVariant:intrinsicnerf} with our own results, which showed our separation in shadow areas were superior to existing methods.

Unlike other methods, our method first reads the 2D instance annotations obtained through semantic segmentation, and performs the intrinsic decomposition by assigning the ``hue'' and ``brightness'' to each instance. It allows the users to control the distribution of shadows in the scene. Compared with other methods, our algorithm has better performance for of plausible and consistent albedo estimation results.

\begin{figure}[htbp]
	\begin{center}
		\includegraphics[width=\linewidth]{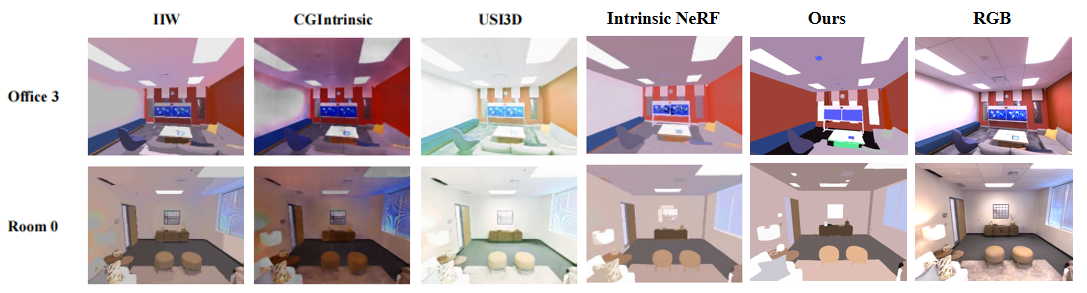}
	\end{center}
	\caption{Illustration of the generated intrinsic decomposition image, compared with other methods. The main method Intrinsic NeRF~\cite{NeRFVariant:intrinsicnerf}, which did not release code and datasets, mentioned the reference of preprocessed Replica indoor scene dataset provided by Semantic NeRF~\cite{zhi2021semantic}. Therefore we used Replica room\_0 and Replica office\_3 dataset given by Semantic NeRF. The decomposition results are compared with those in the paper. As shown in the figure, the albedo image we got from our method performs better in separation of object and scene environment interaction. For example, the pillow in the lower left corner of the room\_0 image, or the lampshade on the left, our method produces more consistent color blocks that eliminate more details.}
	\label{fig:intrinsic_decomposition_compare}
\end{figure}

Quantitatively, we compared the albedo values of interest points from different viewing directions, and the results showed that in our method, the extraction of albedo attributes from different viewing directions was stable and reliable, while the albedo values obtained by the Intrinsic NeRF changed.

\begin{figure}[htbp]
	\begin{center}
		\includegraphics[width=\linewidth]{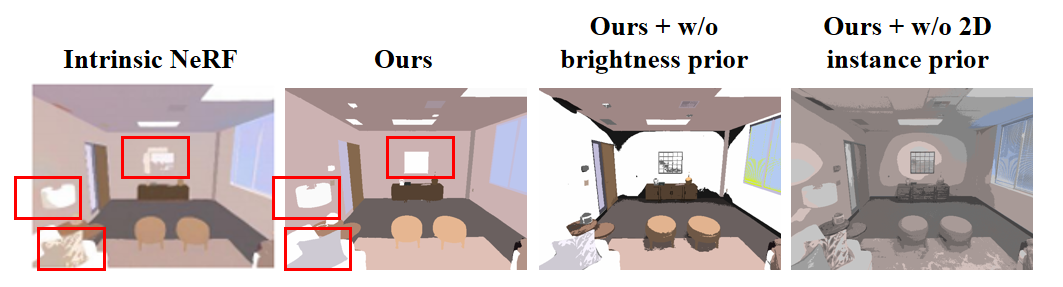}
	\end{center}
	\caption{Some detailed areas, such as shadows. The albedo image obtained by Intrinsic NeRF has shadows and parti-colors that are not considered a good decomposition, as shown in the red box, while our method separate shadows information better. Without \textbf{brightness prior}, parts belonging to the same color could not be recognized, and the darker part result from shadows would be recognized as a different color. Without \textbf{instance prior}, the similar color of different objects will be grouped together, which makes the intrinsice decomposition much more difficult.}
	\label{fig:intrinsic_decomposition_compare_detail}
\end{figure}

Without the brightness prior, we do not process the image into ``hue'' and ``brightness'' before the intrinsic decomposition, which will result in some shadow areas being recognized as different colors of the object, ultimately unable to separate the shadows from albedo image. Without the 2D instance annotation as supervision, objects can not be distinguished from the scene, which may cause the same color to be painted on different objects.

\begin{table}[htbp]
	\begin{center}
		\caption{Intrinsic Decomposition Comparison and Ablation.}
		\label{tab:albedo_image_comparision}
		\setlength{\tabcolsep}{2mm}
		\begin{tabular}{lccc}
			\toprule[1.3pt]
			
			\multicolumn{ 1}{l}{} & \multicolumn{ 3}{c}{Replica-room\_0} \\
			Config & PSNR$\downarrow$ & SSIM$\uparrow$ & LPIPS$\downarrow$ \\
			\hline
			\specialrule{0em}{1pt}{1pt}
			Intrinsic NeRF~\cite{NeRFVariant:intrinsicnerf} & 30.259  & 0.841 & 0.320 \\
			Ours & 30.209 & 0.828 & 0.275 \\
			Ours + w$\backslash$o brightness prior & 27.653 & 0.653 & 0.270 \\
			Ours + w$\backslash$ 2D instance prior & 29.008 & 0.774 & 0.252 \\
			\bottomrule[1.3pt]
		\end{tabular} 
	\end{center}
\end{table}

What we need to perform a better intrinsic decomposition? In fact, according to Land \etal~\cite{IntrinsicDecomposition:Retinex}, there is no fixed criterion for judging the quality of an intrinsic method. It is more of a reflection of human feelings, which is not qualified to become a strict evaluation criterion, and the real objects in the scene will not exhibit the effect of intrinsic decomposition in any circumstance, unless we re-color the objects and perform a extreme fine illumination. Intrinsic NeRF~\cite{NeRFVariant:intrinsicnerf} is evaluated their albedo images according to the similarity of ground truth provided by earlier methods such as PhySG~\cite{Relighting:physg}. This comparison can only screen out the results more in line with their mission's criteria, while our purpose is to better reflect the separation of hue and shadow. Therefore, we did not use the similar evaluation criteria, but simply compare the albedo results to the original image. Just as we have performed intrinsic decomposition based on two assumptions, we divided the image into 2 parts: hue and brightness. We compared the PSNR score and SSIM score between the hue from the albedo image to that from original image, while the LPIPS score between the brightness from the albedo image to that from original image. Tab. (\ref{tab:albedo_image_comparision}) shows our comparison results.

The comparison between hues has no brightness difference, so the closer the colors are, the closer the shadow effects are reflected to the albedo image, which means the shadow separation is not performed well. Therefore, in contrast to common sense, in our evaluation, the lower PSNR score is usually better. However, the PSNR score itself is not enough, because a single white image can provided an even lower score, so we need to combine the SSIM and LPIPS comparison to judge an albedo image's separation effect. The SSIM score~\cite{wang2004SSIM} pays more attention to the local details of the image. We can use this criterion to filter out the ``pure white'' situations, which lost almost all the scene structure information during decomposition process. The higher SSIM score is the better. LPIPS score~\cite{zhang2018LPIPS} shows the relationship between the shadow information we obtained from the albedo image and the original image. Its advantage is that it is closer to human perception, thus the most important shadow information comparison. So as how LPIPS works before, the lower the better. These three criteria can indirectly perform lateral confirmation of our method, that has advantages in the separation performance of scene shadow factor that previous works does not.

\textbf{Analysis of scene inpainting.}

\begin{figure}[htbp]
	\begin{center}
		\includegraphics[width=\linewidth]{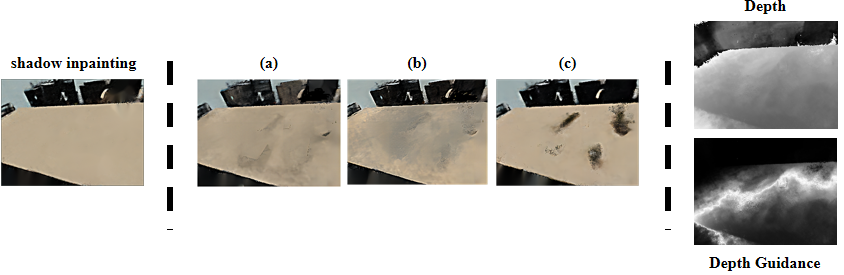}
	\end{center}
	\caption{\textbf{Left:} the result of scene inpainting. \textbf{Middle:} (a) inpainting failure due to incomplete separation of shadows; (b) inpainting failure due to lack of depth guidance; (c) albedo results without scene inpainting step. \textbf{Right:} the depth guidance. For a ``hole pixel'', its ``depth-similar'' pixel region can be drawn as an approximate white line in the scene, indicating the proximity of the albedo colors.}
	\label{fig:shadow_inpainting_ablation}
\end{figure}

To evaluate the effectiveness of our scene inpainting method, we conducted ablation experiments according to the  process of our scene inpainting. Firstly, we need to separate the shadow information from albedo in the intrinsic decomposition step. If it fails, like Fig. (\ref{fig:intrinsic_decomposition_compare_detail}), there will be unremovable shadows in our repaired image, as shown in Fig. (\ref{fig:shadow_inpainting_ablation}) (a).

Secondly, we assign a weight to each albedo pixel, which is equivalent to a ray emitted from camera center to the scene in NeRF volume rendering methods, according to the depth properties of the scene, indicating how similar it is to the albedo colors of the ``hole pixels''. The closer the depths, the higher the similarity of albedo colors. We cancel this step and assign an equal weight to each pixel, as shown in Fig. (\ref{fig:shadow_inpainting_ablation}) (b). The wrong filled colors are combined with other unrelated objects colors.

Finally, we need to set the weights between each of the ``hole pixels'' to 0, preventing their colors from messing with the right filled colors. In Fig. (\ref{fig:shadow_inpainting_ablation}) (c) we show the unfilled results from the dataset. If we edit the objects in such a situation, there will be these unsightly shadows left in the scene, showing the necessity of scene inpainting in our work.

\section{Conclusion and Discussion}

In this paper, we have presented an approach to use intrinsic decomposition method to separate the interaction between objects and scene environment information, to accomplish the editing tasks of the scene arrangement and illumination of the implicit field object without illumination supervision. Firstly, we have discussed the influence of environmental information on 2D images, and then spread this information to 3D spatial properties through the idea of implicit field inverse rendering. To this end, we have provided an OSI-aware system to learn implicit field properties and abstract the influence of this information into an explicit illumination editing component. Our work can freely edit the position orientation of indoor scene objects and their interaction with scene environment, like shadows.

To verify our method, we have tested it on three indoor datasets: Toydesk, Scannet and Replica. The comparison results have show that our method is superior to other implicit field editing works in scene editing task, and is comparable to the existing methods in rendering quality. At present, due to the innate ``absorbing'' light model of implicit field, the information abstraction of our method deviates from the original real light mechanism. Therefore, it can only produce a synthetic shadow editing effects. How to further recognize the real scene illumination information recovery and interaction between objects and scene environment under unsupervised illuminating conditions will be our next research direction.

\bibliographystyle{ACM-Reference-Format}

\bibliography{sample-base}


\end{document}